\documentclass{article}

\PassOptionsToPackage{numbers, compress}{natbib}
 \usepackage[preprint]{neurips_2026}

\usepackage{amsmath,amsfonts}
\usepackage{graphicx}
\usepackage{booktabs}
\usepackage{multirow}
\usepackage{xcolor}

\usepackage[utf8]{inputenc}
\usepackage[T1]{fontenc}
\usepackage{url}
\usepackage{hyperref}

\usepackage{enumitem}

\usepackage{nicefrac}
\usepackage{microtype}
\usepackage{orcidlink}

\usepackage{cleveref}
\usepackage[table]{xcolor}

\usepackage{algorithm}
\usepackage{algorithmic}
\usepackage{pifont}
\usepackage{amssymb}

\newcommand{\xmark}{\raisebox{0.2ex}{\scalebox{0.9}{$\times$}}}
\def\Methodname{AlignDrive}

\title{AlignDrive: Aligned Lateral-Longitudinal Planning for End-to-End Autonomous Driving}

\author{
Yanhao Wu$^{1,2, *}$  ~
Haoyang Zhang$^{2, \ddagger}$ ~
Fei He$^{2}$ ~
Rui Wu$^{2}$ \\
~\textbf{Yanhu Shan}$^{2}$~
~\textbf{Congpei Qiu}$^{1}$~
\textbf{Liang Gao}$^{1}$ ~
\textbf{Wei Ke}$^{1}$ ~
\textbf{Tong Zhang \textdagger}$^{3}$ \\
\\[-1em]
$^1$ School of Software Engineering, XJTU \quad
$^2$ Horizon Robotics \quad  \\
$^3$ University of Chinese Academy of Sciences
}

\begin{document}

\maketitle

\vspace{-2em}
\begin{abstract}
\vspace{-1em}
Practical autonomous driving requires models that generalize by reasoning through spatial-temporal possibilities to exclude unsafe outcomes. While state-of-the-art (SOTA) methods use parallel planning architectures, they fail to explicitly couple speed decisions with agent behavior along the driving path, leading to suboptimal coordination.
To address this, we propose a cascaded framework that transforms longitudinal planning from an independent prediction task into a path-conditioned reasoning process. 
On the model side, we introduce an anchor-based regression design that conditions longitudinal prediction on the lateral drive path, and reformulate longitudinal planning as 1D displacement prediction along the path. This reduces geometric uncertainty and sharpens the model's focus on interaction-driven dynamics. 
On the data side, we introduce a planning-oriented data augmentation strategy that simulates rare safety-critical events by programmatically inserting agents and relabeling longitudinal targets to enforce collision avoidance.
Evaluated on the challenging Bench2Drive benchmark, our method achieves SOTA performance with a driving score of 89.07 and a success rate of 73.18\%, demonstrating significantly improved coordination and safety. Further evaluation on Fail2Drive confirms strong generalization to rare edge cases where parallel formulations typically fail. \noindent{Project page:} \href{https://yanhaowu.github.io/AlignDrive/}{https://yanhaowu.github.io/AlignDrive/}
\footnote{*:~Work done during internship at Horizon Robotics. \textdagger: Corresponding author. $^{\ddagger}$: Project leader.
}
\end{abstract}

\vspace{-0.7cm}
\section{Introduction}
\vspace{-0.4cm}

End-to-end autonomous driving has rapidly advanced, enabling joint perception and planning in increasingly complex environments~\cite{e2e_SparseDrive, e2e_ipad, e2e_drivetransformer, e2e_momad, e2e_uniad, e2e_paradrive}. A key requirement for practical deployment is generalization: the ability to reason through a wide range of spatial-temporal possibilities and systematically exclude unsafe outcomes, particularly in rare, safety-critical scenarios that are underrepresented in standard training data.

\begin{figure}[t]
        \centering    
        \hspace{0.5 cm}
        \includegraphics[width=1\linewidth]{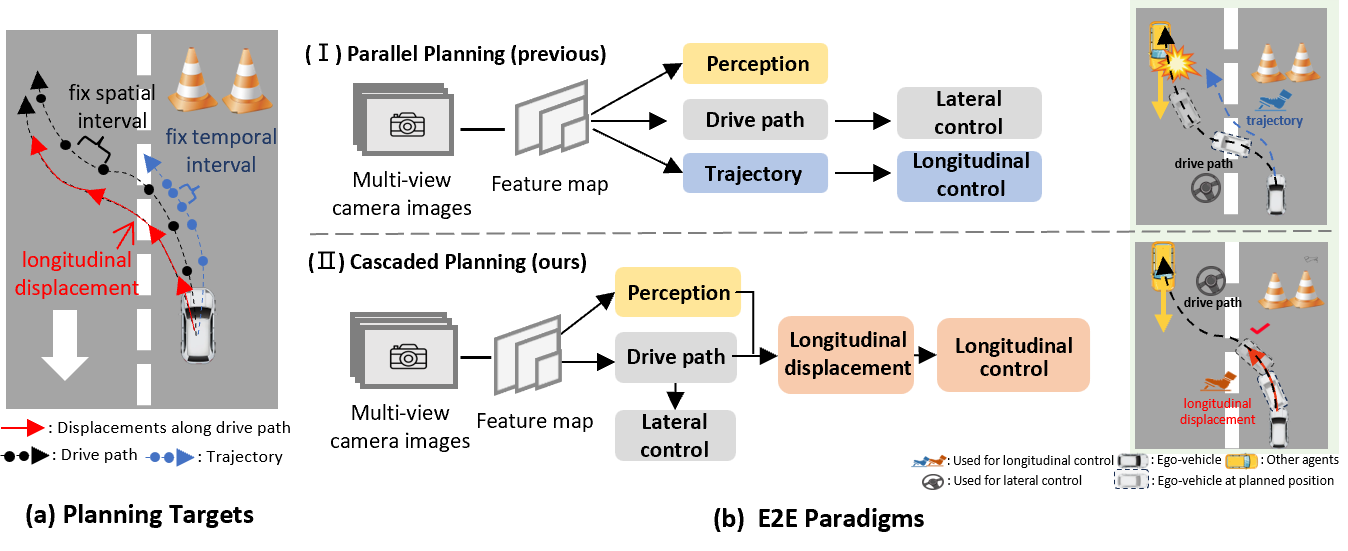} 
        \vspace{-0.8cm}
        \caption{(a) Drive path (black), trajectory (blue), and longitudinal displacement (red). Path waypoints are sampled spatially, trajectory waypoints temporally, and displacements represent traveled distance along the path at fixed time intervals. (b) Comparison of E2E paradigms. 
        {Parallel planning predicts the drive path and longitudinal trajectory independently, which can lead to potential coordination inconsistencies. In the example on the right, the independently predicted longitudinal trajectory is collision-free, but applying its speed along a separately predicted lateral path could cause a collision. In contrast, our cascaded paradigm first predicts the drive path and then regresses path-conditioned longitudinal displacements. With the path prior, the model identifies the potential conflict and outputs shorter displacements, yielding to avoid collision. Perception inputs are omitted for clarity.}         
        \vspace{-0.7cm}}
        \label{fig:teaser}
\end{figure}

A promising direction in recent work is to decouple planning into separate lateral and longitudinal prediction tasks~\citep{e2e_TF++, e2e_carllava, e2e_HiPAD}. In this paradigm, lateral planning predicts a drive path—waypoints sampled at fixed spatial intervals—as the target for steering control, while longitudinal planning predicts a trajectory—waypoints sampled at fixed temporal intervals—as the target for speed control. This disentanglement has proven effective for improving lateral planning performance. However, reliable longitudinal planning remains a major challenge. In particular, SOTA methods such as HiP-AD~\cite{e2e_HiPAD} employ a parallel architecture in which drive path and trajectory queries are decoded by independent heads simultaneously. While this design is effective overall, the parallel formulation does not explicitly couple longitudinal decisions with surrounding agent behavior along the driving path. As a result, even when predicted drive paths are geometrically accurate, the corresponding speed profiles can become unsafe in complex interactive scenarios—for instance, failing to brake in response to a sudden cut-in or a decelerating lead vehicle (see Fig.~\ref{fig:teaser}(b)). 

We argue that reliable longitudinal planning plays an important role in robust driving generalization, as interaction-heavy safety-critical scenarios critically depend on accurate speed reasoning conditioned on surrounding agent behavior. This motivates explicitly aligning longitudinal motion reasoning with the dynamics of surrounding agents along the intended driving path. To this end, we address the problem from two complementary perspectives: model design and data construction. 

On the model side, we propose a cascaded framework that conditions longitudinal planning on the lateral drive path via an anchor-based regression design. Rather than predicting full 2D trajectory waypoints independently, our formulation first establishes the drive path as a structured geometric prior, then predicts longitudinal motion as 1D displacement values along that path (see Fig.~\ref{fig:teaser}(a)). This cascaded design offers two key benefits. First, anchoring longitudinal reasoning to the drive path reduces geometric uncertainty and provides a coarse but explicit alignment between longitudinal motion decisions and the spatial context of surrounding agents. Second, reformulating longitudinal planning as 1D displacement removes unnecessary geometric degrees of freedom, sharpening the model's focus on interaction-driven dynamics and resulting in more collision-aware longitudinal planning (see Fig.~\ref{fig:teaser}(b)).

On the data side, we introduce a planning-oriented data augmentation strategy to improve generalization. By programmatically inserting agents and relabeling 1D displacement targets to enforce collision avoidance, we simulate safety-critical events that are severely underrepresented in real-world logs. Whereas parallel 2D models fail to learn effectively from such augmentation because steering and speed are geometrically entangled, our 1D formulation \textbf{untangles} these variables. This structural clarity allows the model to learn causal collision-avoidance logic rather than merely memorizing expert patterns, enabling robust generalization to unseen interactive edge cases.


Building on these insights, we develop an E2E driving framework~\Methodname~that conditions longitudinal planning on the drive path and leverages this formulation to enable effective data augmentation, with code and models to be publicly released. Overall, our contributions are threefold: 
\vspace{-0.6cm}
\begin{itemize}[leftmargin=30pt]
\item We propose a novel cascaded planning paradigm where longitudinal planning is explicitly conditioned on a predicted lateral drive path. This method establishes a tight coupling between the two tasks, using the path as geometric priors for subsequent longitudinal planning.
\item Based on this paradigm, we reformulate the longitudinal planning task as a simpler 1D displacement prediction problem along the drive path. This allows the model to focus its capacity on crucial dynamic interactions rather than redundantly encoding static geometry.
\item We introduce an effective, planning-oriented data augmentation strategy. By programmatically modifying only the 1D displacement labels in response to inserted agents, we can generate diverse and realistic safety-critical training scenarios that are rare in logged data.  

\vspace{-0.2cm}
\end{itemize}
We primarily evaluate our method on the popular closed-loop benchmark Bench2Drive~\cite{benchmark_bench2drive}, where~\Methodname~demonstrates clear improvements over state-of-the-art techniques. To further verify generalization, we additionally evaluate on the recently proposed Fail2Drive benchmark~\cite{benchmark_fail2drive}, as well as report open-loop results on the nuScenes dataset~\cite{dataset_nuscene}. The results consistently demonstrate the strong performance, generalization ability, and robustness of our method. Together, these experiments provide a comprehensive validation of the model’s capabilities.
\vspace{-0.4cm}

\section{Related Work}
\vspace{-0.3cm}
\subsection{End-to-end Autonomous driving }
\vspace{-0.2cm}
End-to-end autonomous driving methods~\citep{e2e_diffad, e2e_HiPAD, e2e_rad, e2e_SparseDrive, e2e_uniad, e2e_vad, e2e_goalflow} have rapidly advanced in recent years, with trajectory planning playing a central role in predicting the ego vehicle’s future states. One line of work, exemplified by SparseDrive~\citep{e2e_SparseDrive, e2e_momad}, directly predicts trajectories in an end-to-end manner. While effective in nominal scenarios, this joint prediction paradigm struggles to achieve accurate lateral and longitudinal planning simultaneously. TF++~\citep{e2e_TF++} predicts the drive path and instantaneous vehicle speed in parallel, with speed treated as a classification task. However, the coarse discretization of velocity limits planning accuracy. More recent approaches, including HiP-AD~\citep{e2e_HiPAD} and Carllava~\citep{e2e_carllava}, instead decouple path and trajectory prediction. HiP-AD employs independent heads, while Carllava, built on a LLaVA-like architecture~\citep{llava}, generates the drive path and trajectory sequentially as output tokens. However, these methods still rely on predicting full waypoints rather than explicit longitudinal displacements, limiting alignment between longitudinal planning and surrounding agents in challenging scenarios such as sharp turns or dynamic interactions. In contrast, we propose a cascaded, anchor-based formulation that first predicts the drive path and then forecasts a sequence of future longitudinal displacements along it. Our approach regresses offsets from predefined anchors using a two-stage design with dedicated modules for path and displacement prediction, rather than jointly generating them as tokens. This naturally enforces lateral–longitudinal consistency, simplifies reasoning about dynamic interactions, and improves path-following safety.

\vspace{-0.4cm}
\subsection{Data augmentation}
\vspace{-0.3cm}
Data augmentation is widely employed in multiple fields~\citep{da_logo, da_refine, da_STTSL, da_sparse4d}. In autonomous driving, it is commonly applied to augment image data through techniques such as cropping, flipping, and color jittering~\citep{e2e_SparseDrive}, which improve the model's ability to generalize across varying visual conditions and strengthen perception robustness. 
Pluto~\citep{planner_pluto} employs agent drop and insertion as data augmentation strategies to generate both positive and negative scene samples. These augmented samples are utilized in a contrastive learning framework to enhance the model's scene representation capabilities. However, these augmentations primarily affect perception and influence planning only indirectly.
TF++~\citep{e2e_TF++} introduced an auxiliary camera in the simulation environment, which is randomly repositioned at each time step to increase data diversity. This approach relies on additional simulator equipment and focuses primarily on lateral recovery, providing limited guidance for longitudinal planning or dynamic interaction reasoning.
In contrast, our planning-oriented augmentation is directly coupled with longitudinal planning, operating on the percepted agents and adjusting longitudinal displacements. This forces the model to focus explicitly on dynamic agent interactions, enabling path-consistent and collision-aware planning in rare, safety-critical scenarios.
\vspace{-0.3cm}

\section{Method}
\vspace{-0.2cm}
\subsection{Overview}
\begin{figure}[t]
        \centering    
        \includegraphics[width=1\linewidth]{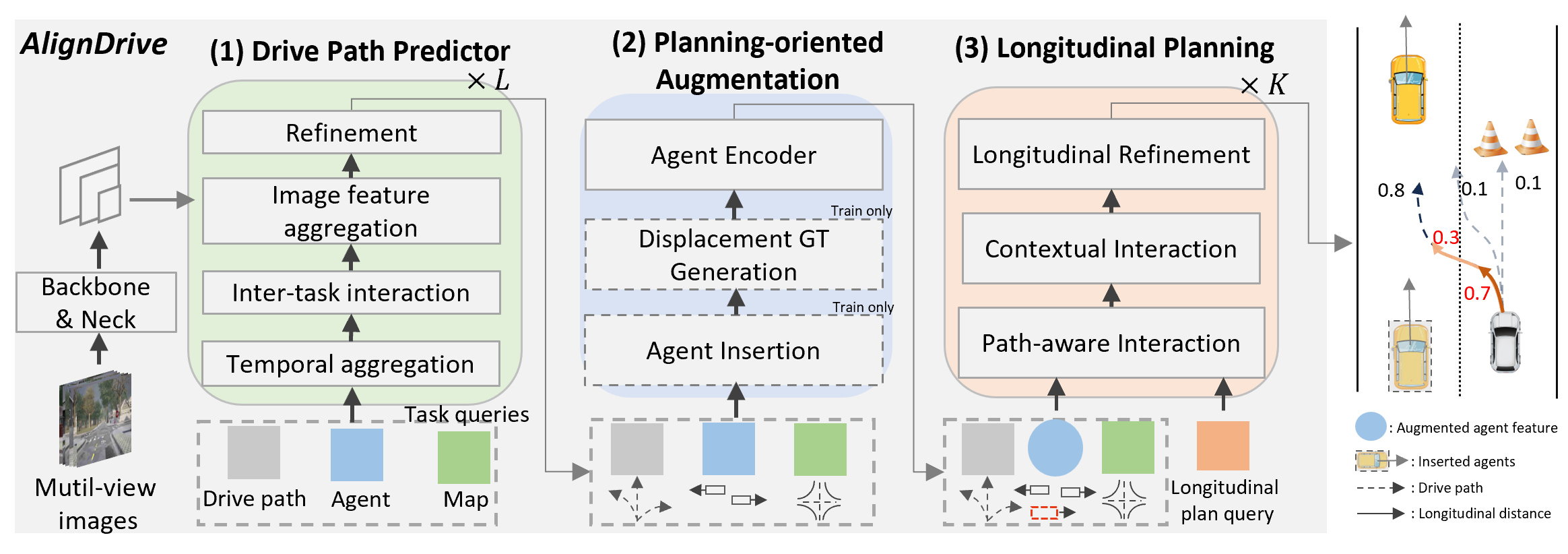} 
        \vspace{-0.8cm}
        \caption{Overview of the proposed \Methodname~system, which consists of three components. The Drive Path Predictor refines queries through cross-attention with image features to encode the drive path, maps, and agents. The Planning-oriented Data Augmentation enriches scenarios by inserting additional agents and relabeling longitudinal displacements. Finally, the Longitudinal Planning Module predicts forward displacements along the drive path; combined with the path, these displacements yield the final trajectory that is both collision-aware and spatially consistent. On the right side of the figure, the black numbers denote the scores of predicted drive paths, while the red numbers represent the scores of the corresponding longitudinal planning for each drive path.}
        
        \label{fig:method}
        \vspace{-0.5cm}
\end{figure}

\vspace{-0.2cm}
Figure~\ref{fig:method} provides an overview of \Methodname, which consists of three main components. 
The \textbf{Drive Path Predictor} refines queries via cross-attention with image features~\citep{da_sparse4d}, producing representations of the drive path, map, and dynamic agents. 
The \textbf{Planning-oriented Data Augmentation} module decodes agent queries into bounding boxes, re-encodes them as structured features, and enables insertion of synthetic agents with relabeled longitudinal displacements for consistent supervision. 
The \textbf{Longitudinal Planning} module then predicts displacements along the drive path from enriched queries, ensuring spatial consistency and collision awareness. 
This design preserves end-to-end training while supporting robust planning. We discuss components below.
\vspace{-0.2cm}
\subsection{Drive Path Predictor}
\vspace{-0.2cm}

Let us denote multi-scale features from $V$ camera views as $\{\mathbf{f}_{i}\}_{i=1}^{V}$.  
Based on training data, we cluster ground-truth annotations to obtain anchors, which differ by task modality: bounding boxes for agents, and typical polylines for map elements and drive paths.
We denote them as $\mathbf{A}_a \in \mathbb{R}^{N_a \times D_a}$, $\mathbf{A}_m \in \mathbb{R}^{N_m \times D_m}$, and $\mathbf{A}_d \in \mathbb{R}^{N_d \times D_d}$, where $D_a, D_m, D_d$ are the dimensions of each type.  
Based on these anchors, we initialize three sets of task queries: agent queries $\mathbf{Q}_a \in \mathbb{R}^{N_a \times C}$, map queries $\mathbf{Q}_m \in \mathbb{R}^{N_m \times C}$, and drive path queries $\mathbf{Q}_d \in \mathbb{R}^{N_d \times C}$, where $C$ is the feature dimension. The Drive Path Predictor consists of $L$ stacked blocks, within which queries interact with image features, historical information, and each other. Through these interactions, the corresponding anchors are iteratively refined across blocks, yielding progressively updated estimates.

\textbf{Temporal Aggregation.}  
To incorporate historical information, each query interacts with retained queries from previous frames via  a top-$k$ strategy: 
\begin{equation}
\mathbf{Q}_d \xleftarrow{} \text{Cross-attention}\left(\mathbf{Q} = \mathbf{Q}_d , \mathbf{K} = \mathbf{Q}_d^{t-T_{p}:t-1}, \mathbf{V} = \mathbf{Q}_d^{t-T_{p}:t-1}\right),
\end{equation}
where $\mathbf{Q}_d$ is the current drive path queries, and $\mathbf{Q}_d^{t-T_{p}:t-1}$ are historical ones. Map queries $\mathbf{Q}_m$ and agent queries $\mathbf{Q}_a$ are updated similarly. 

\textbf{Inter-Task Interaction.}  
We enable interactions among drive path, agent, and map queries through cross-attention, allowing path queries to be contextually aware of agents and maps, and constraining agent behaviors with map information.
\begin{equation}
\begin{aligned}
&\mathbf{Q}_d \xleftarrow{} \text{Cross-attention}\left(\mathbf{Q} = \mathbf{Q}_d, \mathbf{K} = [\mathbf{Q}_a \| \mathbf{Q}_m], \mathbf{V} = [\mathbf{Q}_a \| \mathbf{Q}_m]\right), \\
&\mathbf{Q}_a \xleftarrow{} \text{Cross-attention}\left(\mathbf{Q} = \mathbf{Q}_a, \mathbf{K} = \mathbf{Q}_m, \mathbf{V} = \mathbf{Q}_m\right), 
\end{aligned}
\end{equation}
\vspace{-0.1cm}
where $ [\cdot \| \cdot] $ denotes concatenation along the token dimension.  

\textbf{Image feature aggregation.}  
To fuse image features, anchors are projected onto multi-view images, and their sampled features are aggregated via deformable attention. For drive path queries:
\begin{equation}
\mathbf{Q}_d \xleftarrow{} \text{DA}\left(\mathbf{Q} = \mathbf{Q}_d, \mathbf{K} = \mathcal{P}(\mathbf{A}_d, \{\mathbf{F}_i\}_{i=1}^V), \mathbf{V} = \mathcal{P}(\mathbf{A}_d, \{\mathbf{F}_i\}_{i=1}^V)\right),
\end{equation}
where $\text{DA}$ is deformable attention and $\mathcal{P}(\cdot)$ denotes projection and sampling. Map queries $\mathbf{Q}_m$ and agent queries $\mathbf{Q}_a$ are enhanced in the same way using their anchors $\mathbf{A}_m, \mathbf{A}_a$.  

\textbf{Refinement.}  
The model iteratively refines its predictions across the L blocks. In the refinement stage of each block, for a given anchor $\mathbf{A}_d$, we first generate a feature embedding using a task-specific encoder, $MLP_{enc}(A_d)$. This embedding is fused with the corresponding query $\mathbf{Q}_d$ and fed into an MLP to predict a corrective offset, $\Delta \mathbf{Y}_d$. The anchor is then updated by applying this offset. This process allows the model to progressively improve its estimate from a coarse anchor to a precise prediction.
\begin{equation}
\Delta \mathbf{Y}_d = \text{MLP}\left(\mathbf{Q}_d + \text{MLP}_{enc}(\mathbf{A}_d)\right), \quad 
\mathbf{A}_d \leftarrow \mathbf{A}_d + \Delta \mathbf{Y}_d,
\end{equation}
where $\Delta \mathbf{Y}_d \in \mathbb{R}^{N_d \times D_d}$ is the predicted offset for each drive path anchor, 
and $D_d = P \times 2$ corresponds to $P$ future waypoints. The refined waypoints are obtained as $\hat{\mathbf{Y}}_d = \mathbf{A}_d$ after iteratively updating the anchors through all $L$ blocks. A separate MLP head is applied to the drive path query to predict confidence scores $\mathbf{S}_d \in \mathbb{R}^{N_d \times 1}$ for candidate drive paths.
Map anchors are refined in the same iterative manner using $\mathbf{Q}_m$ and $\mathbf{A}_m$.  

For agents, static attributes (e.g., position, size, orientation) are predicted using an MLP applied to the queries combined with anchor features, while dynamic motion is predicted directly from queries without anchor-based refinement:
\begin{equation}
\hat{\mathbf{Y}}_a^{state} = \text{MLP}_{state}\left(\mathbf{Q}_a + E(\mathbf{A}_a)\right), \quad 
\hat{\mathbf{Y}}_a^{motion} = \text{MLP}_{motion}\left(\mathbf{Q}_a\right),
\end{equation}
where $\hat{\mathbf{Y}}_a^{state} \in \mathbb{R}^{N_a \times S}$ contains $S$ attributes for each agent, and  
$\hat{\mathbf{Y}}_a^{motion} \in \mathbb{R}^{N_a \times T \times 2}$ contains the predicted future trajectories.

\begin{figure}[t]
        \centering    
        \hspace{0.5 cm}
        \includegraphics[width=1\linewidth]{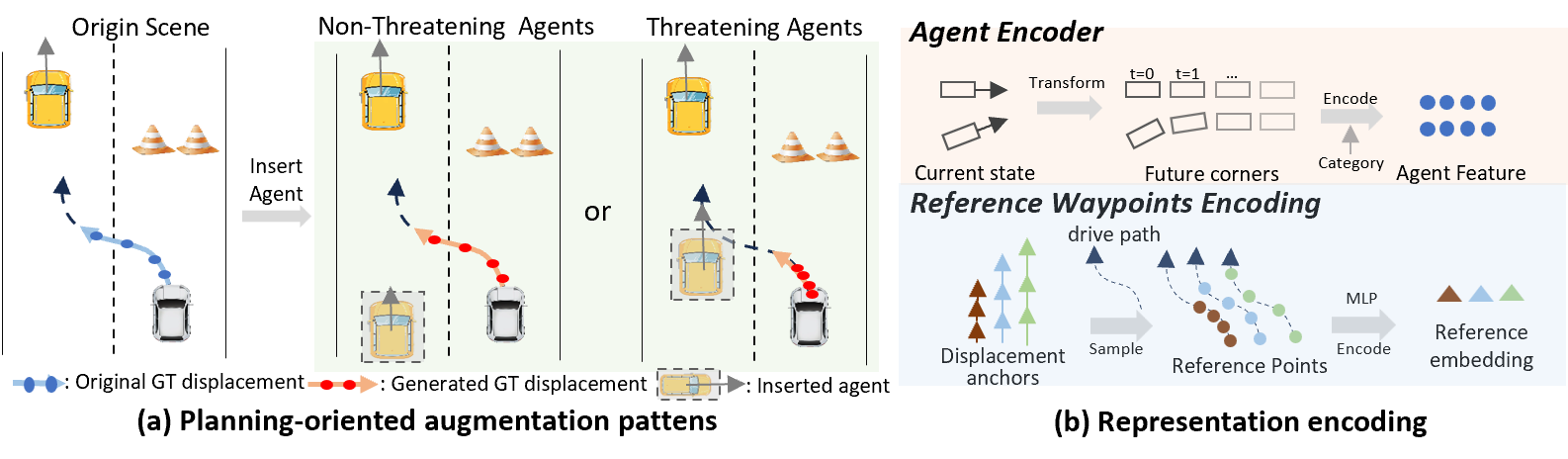} 
        \vspace{-0.9cm}
        \caption{(a) Planning-oriented augmentation. Non-threatening agents are inserted at a distance with unchanged GT displacements, while threatening agents are placed nearby and cause adaptive shortening of GT displacements. 
        (b) Representation encoding. Inserted agents are projected into future positions, transformed to corner representations, and encoded via a Fourier encoder (top). Reference points are sampled by displacement anchors and encoded with MLPs. For clarity, although multiple drive paths are predicted in practice, only one representative path is illustrated here (bottom).} 
        \label{fig:module_detail}
        \vspace{-0.5cm}
\end{figure}

\subsection{Planning-oriented Data Augmentation}
\vspace{-0.2cm}
To enrich interactive scenarios, we insert a virtual agent into the detected agents with probability $\alpha$. 

\textbf{Agent Insertion.}
The virtual agent is initialized with a randomly sampled state $\mathbf{X}_{vir}$ and a target point $\mathbf{P}^*$, selected on the ego vehicle's ground-truth drive path $\mathbf{Y}_d^{\text{GT}}$. Together, they determine the virtual agent's future motion.
As illustrated in Fig.~\ref{fig:module_detail}(a), the virtual agent randomly adopts one of two velocity patterns: gradually approaching from afar, making it a low-risk and safe maneuver, or approaching faster and closer, potentially colliding with the ego vehicle. To maintain the total number of agents, the original agent with the lowest confidence is removed, and the virtual agent is inserted in its place. The resulting augmented set of agent states and motions are denoted as $\tilde{\mathbf{Y}}^{state}$ and $\tilde{\mathbf{Y}}^{motion}$, ensuring exposure to challenging interactions while preserving scene consistency.

\textbf{Displacement Ground-truth Generation.}  
The ground-truth displacement label is defined as a sequence 
$\mathbf{Y}_d^{\text{GT}} \in \mathbb{R}^{T \times 1}$, representing $T$ future longitudinal displacements of the ego vehicle along the drive path, each measured over a fixed temporal interval.  
When a virtual agent is inserted that would collide with the ego vehicle, we first determine the maximum total displacement $D_{\text{safe}}$ the ego vehicle can travel over the $T$ steps without causing a collision.  
Given the original total displacement $D_{\text{orig}} = \sum_{t=1}^T \mathbf{Y}_d^{\text{GT}}[t]$, we compute a safe scaling factor 
$\beta = D_{\text{safe}} / D_{\text{orig}}$ and uniformly scale the original sequence: 
$\tilde{\mathbf{Y}}_d^{\text{GT}} = \beta \cdot \mathbf{Y}_d^{\text{GT}}$.  
This procedure ensures that the displacement at each step is consistently reduced to avoid collisions, while maintaining a dynamically plausible motion profile.

\textbf{Agent Encode.}  
After augmentation, agents are represented using both their states and motions, denoted as $\tilde{\mathbf{Y}}^{\text{state}}$ and $\tilde{\mathbf{Y}}^{\text{motion}}$. These variables are converted into a unified corner-based representation:
\begin{equation}
\mathbf{C}_a = f_{\text{corner}}(\tilde{\mathbf{Y}}^{\text{state}}, \tilde{\mathbf{Y}}^{\text{motion}}), 
\quad \mathbf{C}_a \in \mathbb{R}^{N_a \times (T+1) \times 8},
\end{equation}
where $f_{\text{corner}}(\cdot)$ converts each agent’s bounding box state and future trajectory into the coordinates of its four corners across all time steps, as illustrated in the top of Fig.~\ref{fig:module_detail}(b). To construct agent features, we separately process geometry and category information. The corner representation $\mathbf{C}_a$ is first mapped using Fourier encoder $\Phi$ and passed through $\text{MLP}_{\text{corner}}$~\citep{nerf}, while the agent category (extracted from $\tilde{\mathbf{Y}}^{\text{state}}$) is mapped with $\Phi$ and processed through $\text{MLP}_{\text{cat}}$. Their outputs are then summed to form the unified agent embedding:
\begin{equation}
\mathbf{E}_a^{0:T} = \text{MLP}_{\text{corner}}\!\left(\Phi(\mathbf{C}_a)\right) 
    + \text{MLP}_{\text{cat}}\!\left(\Phi(\text{cat})\right), 
\quad \mathbf{E}_a^{0:T} \in \mathbb{R}^{N_a \times (T+1) \times C}.
\end{equation}
Here, $\mathbf{E}_a^{t} \in \mathbb{R}^{N_a \times C}$ denotes the features of all agents at time step $t$, providing a temporally consistent, category-aware representation. The detailed process can be found in the appendix.

\vspace{-0.2cm}
\subsection{Longitudinal Planning Module}

The longitudinal planning module predicts a sequence of ego vehicle’s future displacements along the drive path at fixed time intervals. It is implemented as $K$ stacked blocks, where queries interact with agent and path features and are progressively refined. We formulate this as an \textit{anchor-based offset regression} task: for each candidate path, $M$ anchors are defined, each representing a sequence of longitudinal displacements for the current step and $T$ future steps, yielding $M \times (T+1)$ learnable queries. 
These queries are responsible for predicting offsets relative to their anchors, thereby coupling the drive path geometry with agent interactions and enabling precise longitudinal planning.
\begin{equation}
\mathbf{A}_l^{0:T} \in \mathbb{R}^{N_d \times M \times (T+1) \times 1}, \quad
\mathbf{Q}_l^{0:T} \in \mathbb{R}^{N_d \times M \times (T+1) \times C},
\end{equation}
where the superscript $0:T$ denotes the stacked sequence of time steps, covering from the current step ($t=0$) to $T$ future steps. Here, $\mathbf{A}_l^{0:T}$ are the anchor displacements and $\mathbf{Q}_l^{0:T}$ their learnable queries. The final displacements are obtained by adding predicted offsets to the anchors.

\textbf{Path-aware Interaction.}  
Each longitudinal planning query is enhanced with reference waypoints sampled along the predicted drive path $\mathbf{\hat{Y}}_d$ at anchor displacements $\mathbf{A}_l^{0:T}$. 
Let sampled points $\mathbf{P}_l$ be
\begin{equation}
\mathbf{P}_l = \text{Interp}(\mathbf{\hat{Y}}_d, \mathbf{A}_l^{0:T}), \quad \mathbf{P}_l \in \mathbb{R}^{N_d \times M \times (T+1) \times 2},
\end{equation}
where $\text{Interp}(\cdot)$ denotes linear interpolation along the path according to cumulative anchor displacements. The points for each anchor are encoded jointly into a feature vector, as shown in Fig.~\ref{fig:module_detail}(b) (bottom):
\begin{equation}
\mathbf{F}_l = \text{MLP}(\mathbf{P}_l), \quad \mathbf{F}_l \in \mathbb{R}^{N_d \times M \times C}.
\end{equation}
Then, this feature is broadcasted across the T+1 time steps and added to the longitudinal query for cross-attention with the drive path queries $\mathbf{Q}_d$:
\begin{equation}
\mathbf{Q}_l^{0:T} \gets \text{CrossAttn}(\mathbf{Q} = \mathbf{Q}_l ^{0:T}+ \mathbf{F}_l, \mathbf{K} = \mathbf{Q}_d, \mathbf{V} = \mathbf{Q}_d).
\end{equation}
In this way, each longitudinal planning query incorporates both the geometry of the drive path and its anchor-based temporal reference.

\textbf{Contextual Interaction.}  
At each time step $t$, longitudinal planning queries interact with dynamic agents via cross-attention on the encoded agent features to capture agent-specific context:
\begin{equation}
{\mathbf{Q}}_l^{t} \gets \text{CrossAttn}\big(\mathbf{Q}=\mathbf{Q}_l^{t},\ \mathbf{K}=\mathbf{E}_a^{t},\ \mathbf{V}=\mathbf{E}_a^{t}\big),\quad t=0,\dots,T.
\end{equation}
The updated queries attend to map queries via cross-attention, incorporating static cues such as stop lines for longitudinal planning:
\begin{equation}
{\mathbf{Q}}_l^{0:T} \gets \text{CrossAttn}\big(\mathbf{Q}={\mathbf{Q}}_l^{0:T},\ \mathbf{K}=\mathbf{Q}_m,\ \mathbf{V}=\mathbf{Q}_m\big).
\end{equation}
Finally, a temporal positional encoding is added to the queries, and causal self-attention is applied along the temporal dimension to enforce consistency:
\begin{equation}
\mathbf{Q}_l^{0:T} \gets 
\text{CausalSelfAttn}\big({\mathbf{Q}}_l^{0:T} + \text{PE}^{0:T}\big).
\end{equation}

\textbf{Longitudinal Refinement.}  
After obtaining the updated longitudinal queries $\mathbf{Q}_l^{0:T}$ from path-aware and contextual interactions, we enhance them using path-aligned reference points $\mathbf{P}_l$, providing spatial grounding for predicting offsets relative to the anchors. Specifically, the reference points $\mathbf{P}_l$ are encoded by an encoder $\text{MLP}_{ref}$ and fused with the queries to predict offsets:
\begin{equation}
\Delta \mathbf{Y}_l = \text{MLP}\Big(\mathbf{Q}_{l}^{0:T} + 
\text{MLP}_{ref}(\mathbf{P}_l)\Big), \quad
\Delta \mathbf{Y}_l \in \mathbb{R}^{N_d \times M \times (T+1) \times 1}.
\end{equation}
The final longitudinal displacements are obtained by adding offsets to the anchors:
\begin{equation}
\hat{\mathbf{Y}}_l = \mathbf{A}_l^{0:T} + \Delta \mathbf{Y}_l,
\quad
\hat{\mathbf{Y}}_l\in \mathbb{R}^{N_d \times M \times (T+1) \times 1}.
\end{equation}
An auxiliary MLP head is applied to the average of $\mathbf{Q}_l^{0:T}$ over the $T+1$ time steps to predict a confidence score $\mathbf{S}_l \in \mathbb{R}^{N_d \times M \times 1}$ for candidate selection.

Our model outputs $N_d$ candidate drive paths $\hat{\mathbf{Y}}_d$ and, for each drive path, $M$ candidate longitudinal displacement sequences $\hat{\mathbf{Y}}_l$. 
A hierarchical selection strategy~\citep{e2e_SparseDrive} chooses the candidate based on confidence scores $\mathbf{S}_l$ and $\mathbf{S}_d$. 
The selected planning is then tracked using PID controllers. 
Full implementation details are provided in the Appendix~\ref{sec:detail}.

\vspace{-0.2cm}
\subsection{Loss Function}
For planning tasks, we adopt a winner-takes-all strategy to determine which predictions are supervised. The winner is defined as the prediction whose corresponding anchor has the minimum $L_2$ distance from the ground truth. Other losses, including online mapping, agent detection, motion forecasting, and auxiliary tasks, follow \citep{e2e_SparseDrive}.
The total loss is the weighted sum of all components. Details are described in the Appendix~\ref{sup_aux}.
\begin{equation}
\begin{aligned}
\mathcal{L} =\;& \lambda_{\text{map}}\mathcal{L}_{\text{map}}
+ \lambda_{\text{det}}\mathcal{L}_{\text{det}}
+ \lambda_{\text{motion}}\mathcal{L}_{\text{motion}} \\
&~~~~~~~~~~+ \lambda_{\text{drivepath}}\mathcal{L}_{\text{drivepath}}
+ \lambda_{\text{plan}}\mathcal{L}_{\text{plan}}
+ \lambda_{\text{aux}}\mathcal{L}_{\text{aux}} .
\end{aligned}
\end{equation}


\section{Experiments}
\vspace{-0.3cm}
\subsection{Dataset and Metrics}
\vspace{-0.1cm}
\textbf{Dataset}. We utilize the Bench2Drive~\citep{benchmark_bench2drive} benchmark for comprehensive evaluation of our model. 
Closed-loop performance is assessed on 220 standardized test routes to ensure a fair and reproducible comparison. 
To further evaluate generalization under distribution shifts, we additionally conduct experiments on the Fail2Drive benchmark~\citep{benchmark_fail2drive}, which introduces paired-route scenarios for testing robustness in unseen conditions. To assess open-loop performance in the real world, we also conduct experiments on the nuScenes dataset~\citep{dataset_nuscene}. The dataset details can be found in Appendix~\ref {dataset_details}

\textbf{Metrics.} Driving Score (DS), Success Rate (SR), Driving Efficiency (DE), and Comfort. In addition, we introduce a Collision Rate metric, defined as the proportion of scenarios involving collisions with dynamic vehicles, to specifically assess the model’s capability in handling interactive environments. For evaluation on Fail2Drive~\citep{benchmark_fail2drive}, we additionally report the Harmonic Mean (HM) of DS and SR, which provides a balanced metric by penalizing imbalanced performance between the two.


\vspace{-0.3cm}
\subsection{Main Results}
\vspace{-0.2cm}
\begin{table*}[tb]\footnotesize
    \caption{{Closed-loop results of planning in Bench2Drive.
    \textbf{Bold} and \underline{underlined} numbers indicate the best performance within different expert groups.}}
    
    \centering
    {
    \begin{tabular}{l cccc}
    \toprule
    \textbf{Method} & 
    \textbf{Driving Score} ($\uparrow$) & 
    \textbf{SR (\%)} ($\uparrow$) & 
    \textbf{Driving Efficiency} ($\uparrow$) & 
    \textbf{Comfort} ($\uparrow$) \\
    \midrule
    \multicolumn{5}{c}{\textbf{Expert: PDM-Lite~\citep{Expert_PDM}}} \\
    \midrule
    SpaceDrive~\citep{spacedrive}  & 78.02 & 55.11 &- &-\\
    SimLingo~\citep{e2e_simlingo} & \underline{86.02} & \underline{67.27} & 
    \underline{259.23} & \underline{33.67} \\
    \midrule
    \multicolumn{5}{c}{\textbf{Expert: Think2Drive~\citep{e2e_think2drive}}} \\
    \midrule
    VAD~\citep{e2e_vad}                    & 42.35 & 15.00 & 157.94 & 46.01   \\
    SparseDrive~\citep{e2e_SparseDrive}    & 44.54 & 16.71 & 170.21 & \textbf{48.63}   \\ 
    DriveTransformer~\citep{e2e_drivetransformer} & 63.46 & 35.01 & 100.64 & 20.78   \\
    Hydra-NeXt~\citep{e2e_hydranext}       & 73.86 & 50.00 & 197.76 & 20.68 \\
    HiP-AD~\citep{e2e_HiPAD}               & 86.77 & 69.09 & 203.12 & 19.36   \\
    \midrule
    \Methodname(Ours)                       & \textbf{89.07} & \textbf{73.18} & \textbf{212.07} & 16.86   \\
    \bottomrule
    \end{tabular}
    }
    \label{tab:bench2dirve1}
    \vspace{-0.5cm}
    
    \end{table*}

\begin{table*}[tb]\footnotesize
\caption{Multi-Ability Results in Bench2Drive.
}
\centering
{
\begin{tabular}{l|c|ccccc} 
\toprule
\multirow{2}{*}{\textbf{Method}} & 
\multicolumn{1}{c}{\textbf{}} & 
\multicolumn{5}{c}{\textbf{Ability} (\%) $\uparrow$} \\ 
\cmidrule(lr){2-7} 
 & \textbf{Mean} & Merging & Overtaking & Emergency Brake & Give Way & Traffic Sign \\ \midrule
UniAD-Base~\citep{e2e_uniad} & 15.55 & 14.10 & 17.78 & 21.67 & 10.00 & 14.21 \\
VAD~\citep{e2e_vad} & 18.07 & 8.11 & 24.44 & 18.64 & 20.00 & 19.15 \\
DriveTransformer-Large~\citep{e2e_drivetransformer} & 38.60 & 17.57 & 35.00 & 48.36 & 40.00 & 52.10 \\
HiP-AD~\citep{e2e_HiPAD} & 65.98 & 50.00 & \textbf{84.44} & \textbf{83.33} & 40.00 & 72.10 \\
\midrule
\Methodname & \textbf{70.06} & \textbf{75.00} & 75.56 & 75.00 & \textbf{50.00} & \textbf{74.74} \\ \bottomrule
\end{tabular}
}
\label{tab:bench2dirve2}
\vspace{-0.3cm}
\end{table*}
\begin{table*}[tb]\footnotesize
\caption{{Closed-loop results on Fail2Drive benchmark. The best results for RGB-only methods are \textbf{bolded}, and the best for Privileged/Multimodal methods are \underline{underlined}.}}
\centering
\setlength{\tabcolsep}{4.5pt} 
{
\begin{tabular}{l ccc@{\hspace{15pt}}c ccc ccc} 
\toprule
\textbf{Method} &
\textbf{RGB} & \textbf{Lidar} & \textbf{Privilege} & \textbf{Language} &
\multicolumn{3}{c}{\textbf{In-Distribution}} &
\multicolumn{3}{c}{\textbf{Generalization}} \\
& & & & &
\textbf{DS} ($\uparrow$) & \textbf{SR} ($\uparrow$) & \textbf{HM} ($\uparrow$) &
\textbf{DS} ($\uparrow$) & \textbf{SR} ($\uparrow$) & \textbf{HM} ($\uparrow$)\\
\midrule

\multicolumn{11}{c}{\textbf{Privileged Methods}} \\
\midrule
PlanT 2.0~\cite{e2e_plant}  & \xmark & \xmark & \checkmark & \xmark
& 87.8 & 85.0 & 86.4 & 73.3 & 58.0 & 64.8 \\
PDMLite-F2D~\cite{Expert_PDM} & \xmark & \xmark & \checkmark & \xmark
& \underline{95.6} & \underline{97.0} & \underline{96.3} & \underline{94.0} & \underline{95.3} & \underline{94.6} \\

\midrule

\multicolumn{11}{c}{\textbf{Multimodal Methods}} \\
\midrule
Orion~\cite{e2e_origin} & \checkmark & \xmark & \xmark &\checkmark
& 53.0 & 52.0 & 52.5 & 51.2 & 46.0 & 48.5 \\
SimLingo~\cite{e2e_simlingo} & \checkmark & \xmark & \xmark & \checkmark
& 82.6 & \underline{79.3} & \underline{80.9} & 71.7 & 55.0 & 62.2 \\
TF++~\cite{e2e_TF++} & \checkmark & \checkmark & \xmark & \xmark
& \underline{83.3} & 78.5 & 80.8 & \underline{75.4} & \underline{61.1} & \underline{67.5} \\

\midrule

\multicolumn{11}{c}{\textbf{RGB-only Methods}} \\
\midrule
TCP~\cite{e2e_tcp} & \checkmark & \xmark & \xmark & \xmark
& 24.7 & 39.1 & 30.3 & 24.5 & 31.4 & 27.5 \\
UniAD~\cite{e2e_uniad} & \checkmark & \xmark & \xmark & \xmark
& 47.5 & 36.3 & 41.2 & 44.0 & 27.6 & 33.9 \\

HiP-AD~\cite{e2e_HiPAD} & \checkmark & \xmark & \xmark & \xmark
& 74.1 & 70.7 & 72.4 & 67.1 & 56.7 & 61.5 \\
\midrule
AlignDrive & \checkmark & \xmark & \xmark & \xmark
& \textbf{76.0} & \textbf{80.0} & \textbf{78.0} & \textbf{71.4} & \textbf{62.0} & \textbf{66.4} \\

\bottomrule
\end{tabular}
}
\label{tab:fail2drive}
\end{table*}

\begin{table*}[tb]\footnotesize
\vspace{-0.6cm}
\caption{Comparison of inference efficiency and driving performance. \Methodname-Small is a lightweight variant with fewer decoder layers. Experiments are conducted on an RTX 3090 GPU.
}
\centering
{
\begin{tabular}{lcccc}
\toprule
\textbf{Method} &\textbf{Parameters}  &\textbf{Latency} &\textbf{Driving Score} &\textbf{Success Rate (\%)}  \\
\midrule
VAD-Base~\citep{e2e_vad} & - & 224.3 ms & 42.35 & 15.00  \\
DriveTransformer-Large~\citep{e2e_drivetransformer} & 646 M & 221.7 ms & 63.46 & 35.01   \\
HiP-AD~\citep{e2e_HiPAD}  & 97.4 M & 138.9 ms & 86.77 & 69.09 \\
\midrule
\Methodname   & 117.2 M & 177.5 ms & \textbf{89.07} & \textbf{73.18} \\
\Methodname-Small & \textbf{83.7 M} & \textbf{124.5 ms} & \underline{87.45} & \underline{71.82}  \\
\bottomrule
\end{tabular}
}
\label{tab:inference_efficiency}
\vspace{-0.6cm}
\end{table*} 
\textbf{Closed-loop Planning Performance.}
As shown in Tab~\ref{tab:bench2dirve1}, our method achieves strong overall performance, with a Driving Score of 89.07 and a Success Rate of 73.18\%, along with the highest Efficiency of 212.07. The Comfort score is 16.86. This is due to challenging scenarios, such as pedestrian crossings and vehicle cut-ins, which occasionally require abrupt braking or steering. Therefore, comparisons of Comfort are most meaningful among methods with similar Success Rates, as such maneuvers are necessary to ensure safe and successful navigation.

We also report the multi-ability scores in Tab~\ref{tab:bench2dirve2}. Our model achieves the highest overall performance, with a significantly superior average score. Most notably, it reaches a Merging score of 75, far surpassing the previous best of 50. Since merging scenarios involve challenging interactions such as consecutive lane changes and cut-ins, the improvement highlights our model’s enhanced capability in handling dynamic interactions and avoiding collisions, directly validating our claim.

\textbf{Generalization under Distribution Shift.}
As shown in Tab~\ref{tab:fail2drive}, we further evaluate our method on the Fail2Drive benchmark to assess its generalization capability, i.e., the ability to reason through unseen spatial-temporal driving scenarios and avoid unsafe outcomes in rare and challenging interactions. Our method achieves competitive performance among RGB-only approaches, with an In-Distribution DS of 76.0 and an SR of 80.0\%, while maintaining strong generalization performance with a DS of 71.4 and SR of 62.0. Notably, it consistently outperforms prior RGB-only methods under the Generalization setting, demonstrating improved robustness in unseen interaction-heavy scenarios.

These results further validate that our approach not only performs competitively in in-distribution settings but also improves the ability to generalize to unseen spatial-temporal configurations and rare safety-critical interactions, which is essential for real-world autonomous driving systems.

\textbf{Inference Efficiency.}
In addition to planning ability, we evaluate inference efficiency in \cref{tab:inference_efficiency}. Our method achieves the best Driving Score and Success Rate while maintaining lower latency than DriveTransformer and VAD. By reducing the number of stacked blocks, we further develop \Methodname-Small, which is smaller and faster than HiP-AD yet still delivers superior performance, striking a better balance between accuracy and efficiency.

\textbf{Open-loop Evaluation.}
We achieve the lowest collision rate in open-loop evaluation, and the full open-loop results are provided in the Appendix~\ref{addtion_exp}.

Overall, our method consistently achieves strong planning performance and improved robustness under distribution shifts across multiple benchmarks, particularly in challenging interaction-heavy driving scenarios that require reliable longitudinal reasoning.
\vspace{-0.3cm}
\subsection{Ablation Study}
\vspace{-0.2cm}

\begin{table*}[tb]\footnotesize
\caption{{Ablation study on \Methodname~components. LP: uses lateral path prediction to condition longitudinal planning;
DP: formulates longitudinal planning as displacement regression along the drive path; DA: applies planning-oriented data augmentation}}
\centering
{
\begin{tabular}{l|ccc|ccc}
\toprule
\textbf{Variant} & LP & DP & DA & Driving Score $\uparrow$ & Success Rate (\%)$\uparrow$ & Collision Rate (\%) $\downarrow$ \\
\midrule
A & & & & 83.21 & 63.18 & 22.7 \\
B & \checkmark & & & 84.85 & 65.45 & 19.5 \\
C & \checkmark & \checkmark & & 85.82 & 66.81 & 16.3 \\
D & \checkmark & & \checkmark & 86.54 & 68.92 & 15.7 \\
E & \checkmark & \checkmark & \checkmark & \textbf{89.07} & \textbf{73.18} & \textbf{11.4} \\
\bottomrule
\end{tabular}
}
\label{tab:module_ablation}
\vspace{-0.3cm}
\end{table*}

In this section, we perform ablation studies to verify the effectiveness of the key components proposed in \Methodname, directly corresponding to our contributions. 

\textbf{Independent vs Path-Conditioned Longitudinal Planning.}  
We compare Variant A, which predicts lateral drive path and longitudinal trajectories in parallel following prior SOTA methods~\citep{e2e_HiPAD}, with Variant C, our proposed approach that predicts longitudinal displacements along the drive path. This cascaded, path-conditioned design couples lateral and longitudinal planning, resulting in more consistent and effective planning.
As shown in Tab.~\ref{tab:module_ablation}, Variant C achieves a higher overall driving score and increases the Success Rate from 63.18\% to 66.81\%, demonstrating the effectiveness of path-conditioned longitudinal planning.
In addition, Variant C reduces the Collision Rate from 22.7\% to 16.3\%, a 28.2\% relative reduction. This improvement supports our claim that allow longitudinal planning condition on the drive path ensure the model to better focus on dynamic interactions, improving collision avoidance in complex scenarios.

\textbf{Displacement vs Waypoint Prediction.}  
We also evaluate Variant B, which predicts trajectory waypoints conditioned on the drive path at discrete future time steps, rather than predicting longitudinal displacements. Although both variants leverage the drive path as a lateral prior, displacements are more directly associated with dynamic interactions, whereas trajectory waypoints embed additional lateral variations that may dilute this focus. Tab.~\ref{tab:module_ablation} shows that Variant C achieves higher Success Rate and lower Collision Rate, demonstrating that our displacement regression along the drive path is not only conceptually simpler but also empirically superior.

\textbf{Planning-Oriented Data Augmentation.}  
We evaluate the effectiveness of our planning-oriented data augmentation, which inserts synthetic traffic participants and adjusts longitudinal labels while keeping lateral paths unchanged. Variant C without augmentation is compared to Variant E with augmentation. As shown in Tab.~\ref{tab:module_ablation}, augmentation improves overall Driving Score from 85.82 to 89.07 and increases the Success Rate, demonstrating the effectiveness of our strategy. In addition, it reduces the Collision Rate from 16.3\% to 11.4\%, highlighting that augmentation helps the model better handle dynamic agents and improve safety.


\textbf{Displacement Formulation Better Fits Augmentation.}  
We investigate how planning-oriented data augmentation interacts with different longitudinal representations. As shown in Tab.~\ref{tab:module_ablation}, applying augmentation to waypoint-based planning (Variant D) improves Driving Score from 84.85 to 86.54 (+1.69), Success Rate from 65.45\% to 68.92\% (+3.47), and reduces Collision Rate from 19.5\% to 15.7\% (-3.8). In contrast, augmentation paired with displacement-based planning (Variant E) boosts Driving Score from 85.82 to 89.07 (+3.25), Success Rate from 66.81\% to 73.18\% (+6.37), and lowers Collision Rate from 16.3\% to 11.4\% (-4.9). These results indicate that longitudinal displacement formulation better leverages augmentation, yielding larger gains in safety-critical scenarios.

Additional qualitative results are provided in the Appendix~\ref{visulization}, while video demonstrations are included in the supporting materials.

\vspace{-0.3cm}
\section{Conclusion}
\vspace{-0.2cm}
We propose \Methodname, a novel cascaded planning paradigm that explicitly aligns longitudinal planning reasoning with surrounding agent dynamics along the intended driving path. From the model design perspective, longitudinal planning is explicitly conditioned on predicted drive paths, where the path geometry serves as a structured prior for speed reasoning. Building on this, we reformulate longitudinal planning as 1D displacement prediction along the drive path, allowing the model to focus on dynamic interactions rather than redundantly encoding static geometry. From the data construction perspective, we further introduce a planning-oriented augmentation strategy that generates diverse safety-critical interaction scenarios to strengthen longitudinal interaction generalization. Extensive evaluations demonstrate that \Methodname~achieves state-of-the-art performance and robust generalization across challenging driving benchmarks.



\newpage
\bibliographystyle{splncs04}
\bibliography{main}






\appendix
\newpage
\vspace{5cm}

\section*{Technical appendices and supplementary material}

We include the following supplementary content:
\begin{itemize}
    \item Additional visualization demos: a set of videos demonstrating the effectiveness of our method (see the included \texttt{index.html} file).
    \item Original simulation results in the CarlaV2 simulator: detailed scores for each scenario (see \verb|AlignDrive_Meraged_Bench2drive_Results.json|).
\end{itemize}
We include the following content in the appendix:

\begin{itemize}
    \item Additional experiments, including further ablation studies.
    \item Detailed model designs, with full training parameters and algorithmic specifications.
\end{itemize}

\section{IMPLEMENTATION DETAILS}
\label{sec:detail}
\vspace{-0.1cm}
\subsection{Implementation Details}
\vspace{-0.1cm}
The model employs 900 agent queries, 100 map queries, 6 drive path queries, and 5 longitudinal queries. The supervision signal for the drive path is derived from the ego vehicle's ground truth trajectory, sampled at 2-meter intervals. For longitudinal planning, the ground truth is defined as the displacements traveled along the trajectory at a 5Hz sampling rate. The longitudinal planning module employs five constant displacement anchors along the drive path, positioned at 0.25, 1.7, 4.0, 6.0, and 8.5 meters ahead of the current vehicle position. These anchors serve as reference points for predicting future longitudinal displacements. In practice, we set $\alpha = 0.1$, meaning that a virtual agent is inserted in 10\% of training samples. 
\vspace{-0.1cm}
\subsection{Dataset Details}
\label{dataset_details}
\vspace{-0.1cm}
Bench2Drive~\citep{benchmark_bench2drive}.
This dataset consists of 1000 short video clips uniformly sampled from 44 interactive scenarios in CARLA v2~\citep{simulator_carla}. Following the official split, we use 950 clips for training and 50 for validation. Closed-loop performance is evaluated on 220 standardized test routes to ensure fair and reproducible comparison.

nuScenes~\citep{dataset_nuscene}.
nuScenes is a large-scale real-world autonomous driving dataset consisting of 1000 scenes, which are split into 700 training, 150 validation, and 150 test scenes. Each scene contains multi-sensor data collected in complex urban driving environments.

Fail2Drive~\citep{benchmark_fail2drive}.
Fail2Drive does not provide a training set. Following the standard protocol, we directly evaluate using models trained on Bench2Drive. Specifically, we use the weights trained on the Bench2Drive training split without further finetuning. This setup ensures a fair assessment of generalization ability under distribution shifts and unseen paired-route scenarios.

\subsection{Training Details}
During training, the Longitudinal Planning module is initially frozen while the Drive Path predictor is trained for 12 epochs. The Longitudinal Planning module is then unfrozen, and the entire system is trained jointly, with the full training process spanning 36 epochs. Training is conducted on 64 NVIDIA RTX L20 GPUs with a total batch size of 512. We use the AdamW optimizer with weight decay and set the initial learning rate to $1 \times 10^{-4}$. Planning-oriented data augmentation is introduced after 24 epochs to enrich interactive scenarios with virtual agents.

Our model predicts the next $T=15$ drive path waypoints $\{\mathbf{\hat{Y}}^{t}_d\}_{t=1}^{T}$ at 2-meter intervals and longitudinal displacements $\{\mathbf{\hat{Y}}^{t}_l\}_{t=1}^{T}$ at 5 Hz. Supervision is applied using a weighted L1 loss:
\begin{align}
\mathcal{L}_{\text{drivepath}} &= \sum_{t=1}^{T} w_t^{\text{DP}} \|\mathbf{\hat{Y}}^{t}_d - \mathbf{Y}^{t}_d\|_1, \\
\mathcal{L}_{\text{plan}} &= \sum_{t=1}^{T} w_t^{\text{long}} |\mathbf{\hat{Y}}^{t}_l - \mathbf{Y}^{t}_l|,
\end{align}
where the weights assign higher importance to more critical predictions. For the Drive Path waypoints, closer points receive larger weights: $w_t^{\text{DP}} = 1.0$ for $t=1\!-\!5$, $0.6$ for $t=6\!-\!11$, and $0.4$ for $t=12\!-\!15$. A similar time-based weighting $w_t^{\text{long}}$ is applied to longitudinal displacements. This design encourages the model to prioritize predictions that are most critical for immediate planning and safe driving.

The weights for each component of the training objective are set as follows: $\lambda_{\text{map}} = 1$, $\lambda_{\text{det}} = 1$,  $\lambda_{\text{motion}} = 1$, $\lambda_{\text{drivepath}} = 2$, $\lambda_{\text{plan}} = 2$, and $\lambda_{\text{aux}} = 1$.
\vspace{-0.2cm}
\subsection{Model Architecture}
We implement our model using a ResNet-50 backbone ~\citep{resnet} with an input image size of $640 \times 352$. Target waypoints and high-level commands are encoded into plan queries via an MLP. During training, noise is injected into the target waypoints and commands with a certain probability to improve robustness. In the standard version of our model, we employ L = 6 layers in the Drive Path Predictor and K = 6 layers in the Longitudinal Planning module. For the \Methodname-Small variant, we use L = 4 and K = 3.

\vspace{-0.25cm}
\subsection{Planning-oriented data augmentation}
\begin{figure}[t]
\centering
\includegraphics[width=1.0\linewidth]{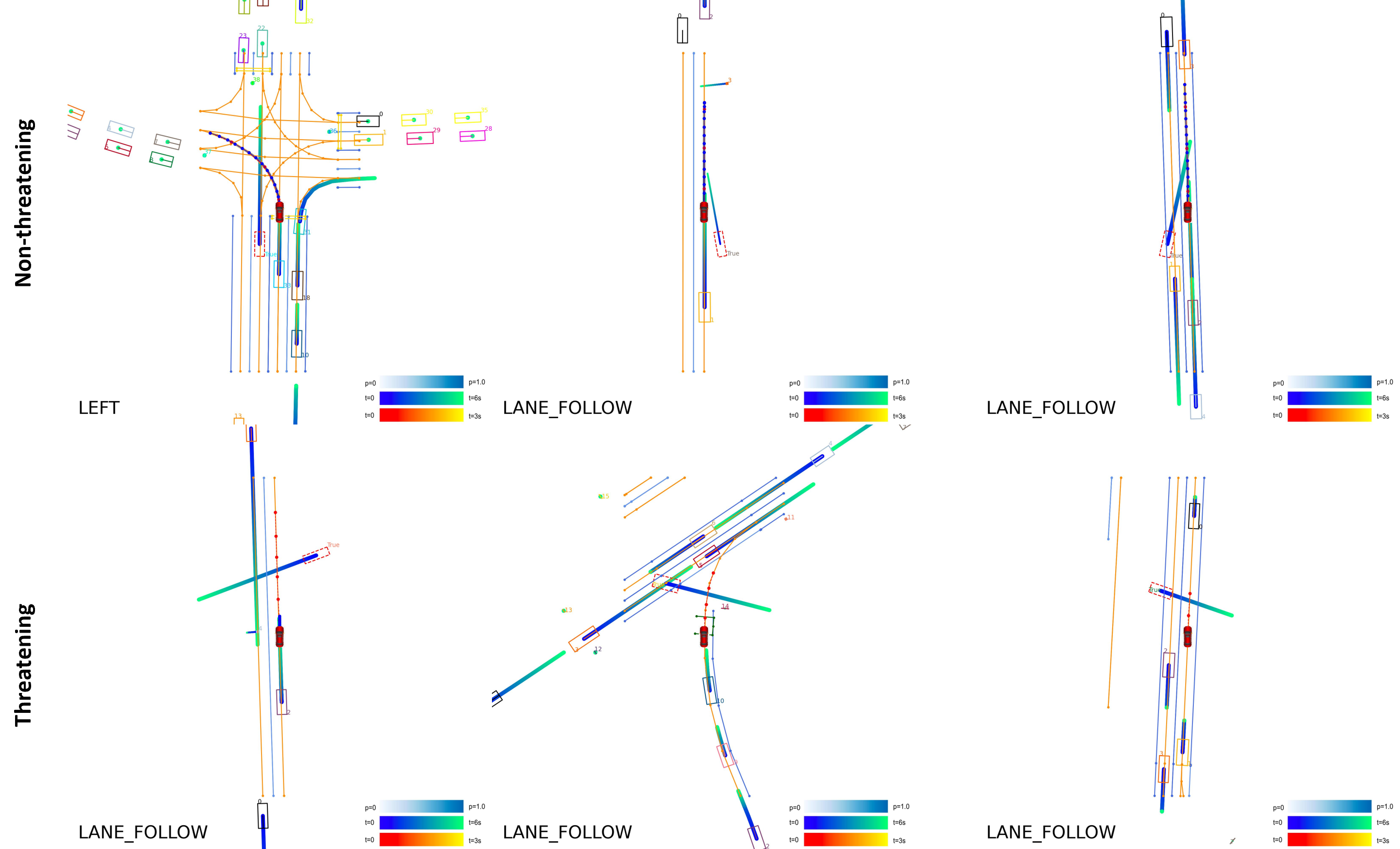} 
\vspace{-0.3cm}
\caption{Visualization of planning-oriented data augmentation. The top row shows non-threatening agents, while the bottom row shows threatening agents. Inserted synthetic agents are indicated with dashed boxes. Red points denote the ego vehicle's original trajectory, and blue lines represent the adjusted longitudinal displacements after augmentation.}
\label{fig:augmentation_show}
\vspace{-0.5cm}
\end{figure}

\vspace{-0.15cm}

\paragraph{Agent Insertion.}  
Our planning-oriented data augmentation begins with the insertion of synthetic agents (see Algorithm~\ref{alg:agent_insertion}). For each training frame, we first compute the ego vehicle's displacement over the next 3 seconds. If this displacement is below a predefined threshold $\delta$, indicating that the ego vehicle is effectively stationary or moving very slowly, no augmentation is performed for that frame (line 10 of Algorithm~\ref{alg:agent_insertion}).

For frames satisfying the displacement criterion, synthetic agents are inserted via a two-step process: selecting an initial position and generating a trajectory. The initial position depends on the agent type: threatening agents are sampled near the ego vehicle, while non-threatening agents are sampled from a distant range. Trajectories are determined by three parameters: the starting position $\mathbf{p}_{start}$, a waypoint $\mathbf{w}$ along the ego vehicle’s future drive path, and the arrival time $t_{arrival}$ at the waypoint. Assuming constant velocity, the agent’s position at each timestep is computed along the straight-line path connecting the start and waypoint. For threatening agents, the arrival time is chosen to potentially induce a collision, whereas non-threatening agents have arrival times that avoid interference. This formulation allows continuous modeling of interactions between the ego vehicle and synthetic agents.

\begin{algorithm}
\caption{Planning-Oriented Agent Insertion}
\label{alg:agent_insertion}
\begin{algorithmic}[1]  

\STATE \textbf{Input:} 
\STATE \quad $\mathbf{Y}_{ego}^{Traj}$: Ego vehicle future trajectory
\STATE \quad $\mathbf{Y}_{ego}^{DrivePath}$: Ego vehicle future trajectory
\STATE \quad $\delta$: Displacement threshold
\STATE \quad $\alpha$: Insertion probability

\STATE \textbf{Output:} 
\STATE \quad $\hat{\mathbf{Y}}_a^{motion}$: Synthetic agent trajectories

\FOR{each training frame}
    \STATE $D \gets \text{ComputeEgoDisplacement}(\mathbf{Y}_{ego}^{future}, 3\text{s})$
    \IF{$D < \delta$}
        \STATE \textbf{continue} 
    \ENDIF

    \IF{$\text{Random}(0,1) \leq \alpha$}
        \STATE $agentRole \gets \text{SelectAgentRole}()$
        
        \IF{$agentRole = \text{threatening}$}
            \STATE $\mathbf{p}_{start} \gets \text{SampleNearPosition}(\mathbf{Y}_{ego}^{Traj})$
        \ELSE
            \STATE $\mathbf{p}_{start} \gets \text{SampleFarPosition}()$
        \ENDIF
        
        \STATE $\mathbf{w} \gets \text{SelectWaypoint}(\mathbf{Y}_{ego}^{DrveiPath})$
        \STATE $t_{arrival} \gets \text{SampleArrivalTime}()$
        \STATE $\hat{\mathbf{Y}}_a^{motion} \gets \text{GenerateTrajectory}(\mathbf{p}_{start}, \mathbf{w}, t_{arrival})$
    \ENDIF
\ENDFOR

\end{algorithmic}
\end{algorithm}

\vspace{-0.15cm}
\paragraph{Displacement Ground-truth Generation.}
With the synthetic agent trajectory inserted, we adjust the ego vehicle’s longitudinal ground-truth displacements to ensure safety (see Algorithm~\ref{alg:gt_generation}). Specifically, we measure the distance between the ego’s predicted positions and the agent at each future timestep within 3 seconds. The last point that satisfies the minimum safety distance is chosen as the new terminal point. We then compute a scaling factor as the ratio between the ego’s travel distance to the adjusted terminal point and the distance to the original terminal point. This factor is used to proportionally shrink the longitudinal displacements between consecutive waypoints, preserving trajectory smoothness while guaranteeing collision-free behavior. The effectiveness of this relabeling procedure is demonstrated in our ablation results (see~\cref{tab:module_ablation}).

\begin{algorithm}[t]
\caption{Displacement Ground-truth Generation}
\label{alg:gt_generation}

\begin{algorithmic}[1]
\STATE \textbf{Input:} Ego future trajectory $\mathbf{Y}_{ego}^{future}$, 
synthetic agent trajectory $\hat{\mathbf{Y}}_a$, 
minimum safe distance $d_{safe}$
\STATE \textbf{Output:} Adjusted ego trajectory $\mathbf{Y}_{ego}^{adjusted}$

\STATE Determine all future timesteps where ego is at least $d_{safe}$ away from the inserted agent
\STATE Let $t_{new}$ be the last safe timestep
\STATE Set $P_{new}$ as the ego position at $t_{new}$ (new 3s terminal point)
\STATE Compute scaling factor $s = $ (distance from start to $P_{new}$) / (distance from start to original terminal point)
\FOR{each consecutive pair of ego future trajectory points}
    \STATE Scale the longitudinal displacement between the points by $s$
\ENDFOR
\STATE $\mathbf{Y}_{ego}^{adjusted} \gets$ updated ego trajectory with scaled displacements
\end{algorithmic}
\end{algorithm}

Figure~\ref{fig:augmentation_show} illustrates our planning-oriented data augmentation. The top row presents non-threatening agents, while the bottom row shows threatening agents. Inserted synthetic agents are highlighted with dashed boxes. Red points indicate the ego vehicle's original trajectory, and blue lines show the adjusted longitudinal displacements after augmentation. 

It is worth noting that our agent insertion relies on minimal rule-based constraints and does not explicitly use road information. As a result, the trajectories of inserted agents may violate road rules. However, this does not negatively impact the longitudinal planning module, which primarily learns to reason about potential interactions with dynamic objects rather than strict road compliance. During training, further constraining inserted agents according to road elements represents a natural extension and a promising direction for future exploration.
\begin{table}[t]
\centering
\footnotesize
\caption{Dataset-specific hyperparameters for nuScenes and Bench2Drive.}
\vspace{-0.1cm}
\label{tab:hyperparameters}
\begin{tabular}{lcc}
\toprule
\textbf{Hyperparameter} & \textbf{nuScenes} & \textbf{Bench2Drive} \\
\midrule
Displacement threshold $\delta$ & 0.5 m & 0.5 m \\
Insertion probability $\alpha$ & 0.05 & 0.1 \\
Safe distance $d_{safe}$ & 3.0 m & 3.0 m \\
Ego trajectory horizon & 3 s & 3 s \\
\bottomrule
\end{tabular}
\end{table} The hyperparameters used in our method are summarized in Table~\ref{tab:hyperparameters}, and we will release our code to facilitate reproducibility.
\vspace{-0.3cm}
\subsection{Auxiliary Tasks}
\label{sup_aux}
We employ two primary auxiliary tasks to improve model learning. 
The first is ego-status prediction. Specifically, an MLP is used to predict the current ego-status of the vehicle from the plan queries, and supervision is applied using an L2 loss.
The second task is inspired by the multi-granularity waypoint prediction used in HiP-AD~\citep{e2e_HiPAD}. In the Drive Path Predictor, we introduce three additional types of queries that interact with the perceived environment in parallel with the drive path query. An Align-fusion strategy~\citep{e2e_HiPAD} is then applied, followed by separate heads to predict: (i) spatial waypoints at 5-meter intervals, (ii) temporal waypoints at 5Hz, and (iii) temporal waypoints at 2Hz. Each prediction is supervised independently. These auxiliary predictions are used only during training and do not participate in inference.


\vspace{-0.2cm}
\subsection{Selection and Control}
\textbf{Selection} The framework produces $N_d$ candidate drive paths and, for each drive path, $M$ longitudinal displacement sequences, representing $N_d \times M$ multimodal predictions that capture both lateral and longitudinal variations. 
First, the drive path with the highest confidence score $\mathbf{S}_d$ predicted by the Drive Path Predictor is selected, along with its corresponding longitudinal displacement candidates $\hat{\mathbf{Y_{l}}}' \in \mathbb{R}^{M \times (T+1) \times 1}$. 
These candidates are further scored $\mathbf{S}_l$, penalizing those that would lead to collisions with predicted motions of other agents, following SparseDrive~\citep{e2e_SparseDrive}. 
The candidate with the highest adjusted score is then chosen as the final output for downstream control.
Importantly, we apply the same strategy to all variants to ensure a fair comparison in ablation studies.

\textbf{Control.}
The selected candidates are executed using two independent PID controllers: one for steering and one for speed. 
The steering controller computes the desired heading based on the selected drive path, while the speed controller computes the desired velocity from the longitudinal displacements. 
Control signals for the vehicle—throttle, brake, and steering angle—are then calculated based on the difference between the desired and the current vehicle states.


\begin{table*}[t]
\centering
\caption{Open-loop planning evaluation results on the nuScenes validation dataset.}
\setlength{\tabcolsep}{4pt}
\scalebox{0.85}{
\begin{tabular}{l|c|cccc|cccc}
\toprule[1pt]
\multirow{2}{*}{\textbf{Method}} &
\multirow{2}{*}{\textbf{Params}} &
\multicolumn{4}{c|}{\textbf{L2 (\textit{m}) $\downarrow$}} &
\multicolumn{4}{c}{\textbf{Collision (\%)} $\downarrow$} \\

& & 1\textit{s} & 2\textit{s} & 3\textit{s} & Avg. 
  & 1\textit{s} & 2\textit{s} & 3\textit{s} & Avg. \\ 
\midrule

VAD-Base~\citep{e2e_vad} & -- & 0.41 & 0.70 & 1.05 & \cellcolor[HTML]{DADADA}{0.72} & 0.03 & 0.19 & 0.43 & \cellcolor[HTML]{DADADA}{0.21} \\
GenAD~\citep{e2e_genad} & -- & 0.28 & 0.49 & 0.78 & \cellcolor[HTML]{DADADA}{0.52} & 0.08 & 0.14 & 0.34 & \cellcolor[HTML]{DADADA}{0.19} \\ 
SparseDrive-S~\citep{e2e_SparseDrive} & 85 M & 0.29 & 0.58 & 0.96 & \cellcolor[HTML]{DADADA}{0.61} & 0.01 & 0.05 & 0.18 & \cellcolor[HTML]{DADADA}{0.08} \\ 
DriveTransformer-Large~\citep{e2e_drivetransformer} & 646 M & 0.16 & 0.30 & 0.55 & \cellcolor[HTML]{DADADA}{\textbf{0.33}} & 0.01 & 0.06 & 0.15 & \cellcolor[HTML]{DADADA}{0.07} \\ 
HiP-AD~\citep{e2e_HiPAD} & 90 M & 0.28 & 0.53 & 0.87 & \cellcolor[HTML]{DADADA}{0.56} & 0.01 & 0.05 & 0.15 & \cellcolor[HTML]{DADADA}{0.07} \\
\midrule 
\Methodname & 147 M & 0.31 & 0.56 & 0.93 & \cellcolor[HTML]{DADADA}{0.60} & 0.00 & 0.02 & 0.17 & \cellcolor[HTML]{DADADA}{\textbf{0.06}} \\
\bottomrule
\end{tabular}
}
\label{tab:nuscenes1_l2_collision}
\end{table*}


\begin{table}[h]\footnotesize
\caption{
Ablation on longitudinal planning (LP), agent query decoding–re-encoding (RE), and planning-oriented data augmentation (DA). 
Decouple: no LP; LP + Original: LP with original agent queries; LP + Reencode: LP with decoded–re-encoded queries; Full (\Methodname): LP + Reencode + DA.
}
\centering
\scalebox{0.9}{
\begin{tabular}{l|ccc|ccc}
\toprule
\textbf{Method} & \textbf{LP} & \textbf{RE} & \textbf{DA} & \textbf{Driving Score} $\uparrow$ & \textbf{Success Rate (\%)} $\uparrow$ & \textbf{Collision Rate (\%)} $\downarrow$ \\
\midrule
Decouple & &  &  & 83.21 & 63.18 & 22.7 \\
LP + Original & \checkmark &  &  & 87.47 & 68.18 & 15.4 \\
LP + Reencode & \checkmark & \checkmark &  & 85.82 & 66.81 & 16.3 \\
Full (\Methodname) & \checkmark & \checkmark & \checkmark & \textbf{89.07} & \textbf{73.18} & \textbf{11.4} \\
\bottomrule
\end{tabular}
}
\label{tab:decode_encode_ablation}
\end{table}
\vspace{-0.1cm}

\begin{table}[t]\footnotesize
\caption{Multiple simulation runs of \Methodname~on Bench2Drive benchmarks. Driving Score, Success Rate, Driving Efficiency, and Comfort are reported for each run along with the average.}
\centering
\scalebox{0.85}{
\begin{tabular}{lcccc}
\toprule
\textbf{Run} & \textbf{Driving Score} $\uparrow$ & \textbf{Success Rate (\%)} $\uparrow$ & \textbf{Driving Efficiency} $\uparrow$ & \textbf{Comfort} $\uparrow$ \\
\midrule
Run 1 & 89.07 & 73.18 & 212.07 & 16.86 \\
Run 2 & 87.80 & 71.36 & 207.85 & 15.25 \\
Run 3 & 88.05 & 70.00 & 210.08 & 17.10 \\
\midrule
\textbf{Average} & 88.30 & 71.50 & 210.00 & 16.40 \\
\bottomrule
\end{tabular}
}
\label{tab:multi_run_ourmethod}
\vspace{-0.2cm}
\end{table}

\vspace{-0.3cm}
\section{More Experiments}
\label{addtion_exp}
\vspace{-1em}
\paragraph{Open-loop results on nuScene} 
\label{open_loop}
We report the open-loop results in ~\cref{tab:nuscenes1_l2_collision}. Our method achieves the lowest collision rate, indicating stronger capability in handling dynamic interactions. Although the L2 distance is not the best, this is influenced by our data augmentation strategy, where inserting additional agents and adjusting the corresponding ground-truth trajectory can introduce discrepancies under an L2-based metric, while other approaches are more directly aligned with such supervision. 

Notably, the very L2 of DriveTransformer-Large~\citep{e2e_drivetransformer} is likely related to its substantially larger model capacity, as increased parameter scale can improve trajectory fitting ability under open-loop supervision. However, such improvements may not directly translate to better real-world planning performance.

Prior work has also noted that open-loop metrics may not fully reflect planning quality due to issues like distribution shift and causal confusion~\citep{openloop_limitaion_1, openloop_limitaion_2, openloop_limitaion_3}. Consistent with this, our method achieves SOTA performance in the closed-loop CARLA evaluation, which offers a more faithful measure of real-world driving behavior. 

\vspace{-0.15cm}
\paragraph{Effect of Re-encoding Agent Queries.} 
Our planning-oriented augmentation requires agent queries to be decoded into bounding boxes and then re-encoded as structured features, which enables the insertion of synthetic agents. This design differs from directly attending to the original agent queries in the longitudinal planning (LP) module, and could potentially affect performance. To further disentangle these factors, we compare four variants: (i) \textbf{Decouple}, which excludes LP and predicts lateral and longitudinal trajectories independently; (ii) \textbf{No-Reencode}, which introduces LP but directly attends to original agent queries without decoding and re-encoding; (iii) \textbf{Reencode}, which uses LP with decoded–re-encoded agent features but without augmentation; and (iv) \textbf{Full} (\Methodname), which combines LP, re-encoding, and planning-oriented augmentation. 

As shown in Tab.~\ref{tab:decode_encode_ablation}, introducing LP (No-Reencode) already improves Driving Score and Success Rate over Decouple, demonstrating that conditioning longitudinal planning on the drive path is effective. Comparing No-Reencode and Reencode reveals a trade-off: directly using original agent queries yields stronger immediate interactions with dynamic agents, but re-encoding is necessary to support augmentation. With augmentation enabled, the Full model achieves the best performance, reducing collision rate most significantly, which confirms that data augmentation and displacement-based LP complement each other in improving robustness, particularly in safety-critical scenarios.

\begin{figure}[t]
\centering
\includegraphics[width=1.0\linewidth]{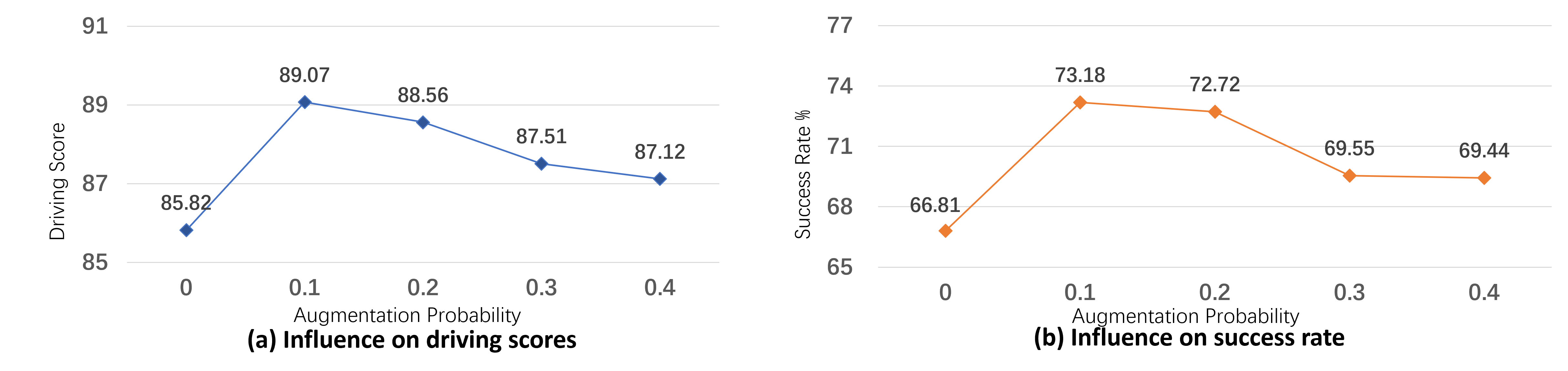} 
\vspace{-0.5cm}
\caption{
Effect of planning-oriented data augmentation on planning performance. All augmented variants ($p=0.1,0.2,0.3,0.4$) outperform the no-augmentation baseline.
}
\vspace{-0.3cm}
\label{fig:ablation_aug}
\end{figure}

\begin{figure}[t]
\centering
\includegraphics[width=1.0\linewidth]{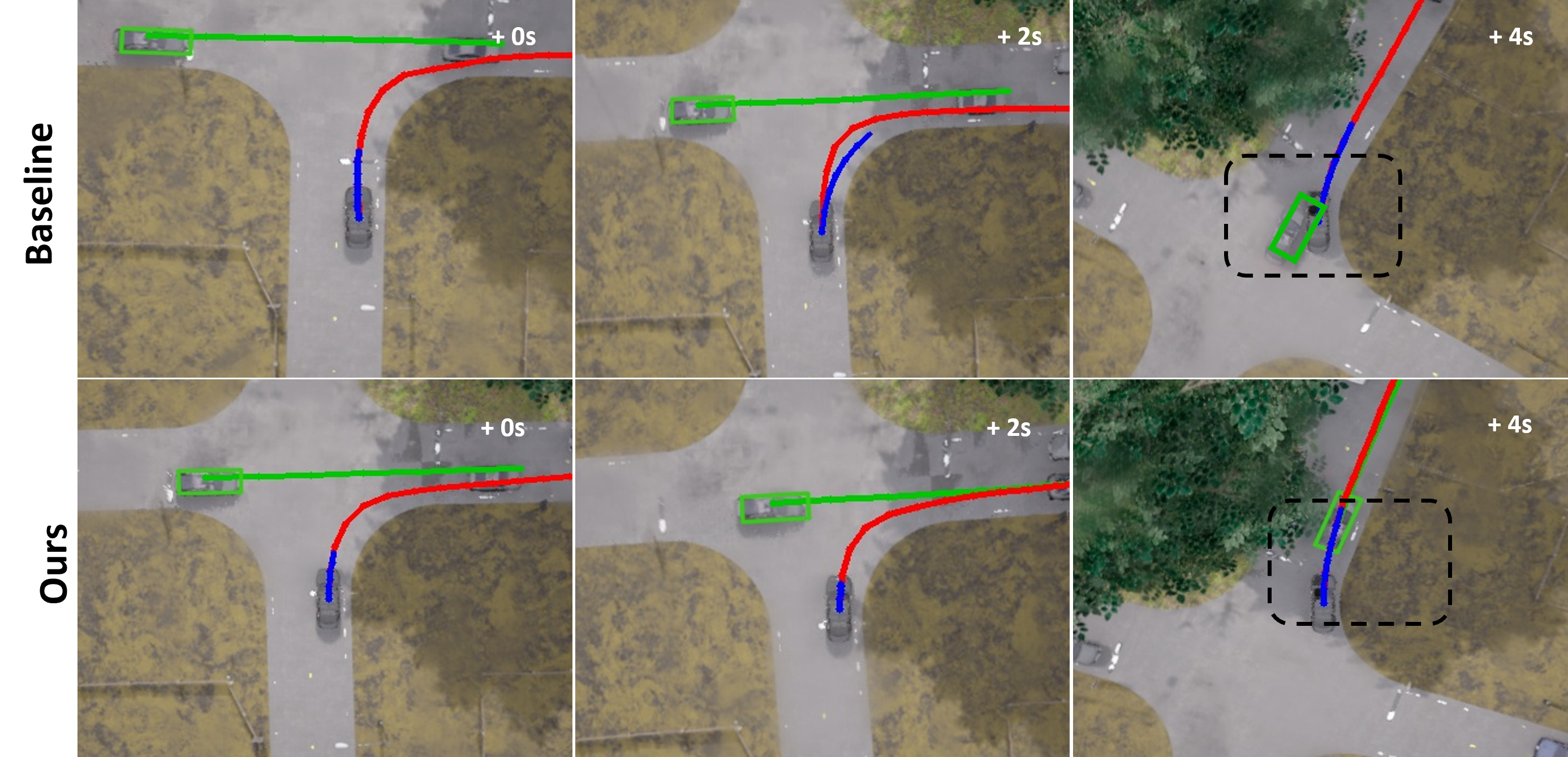} 
\vspace{-0.6cm}
\caption{ Red points are predicted drive paths, while blue points show longitudinal planning outputs (trajectory waypoints for the baseline, displacement sequences for ours). Relevant vehicles are highlighted in green. The baseline collides with cross-traffic, while our method avoids it. }
\label{fig:collision_show}
\vspace{-0.6cm}
\end{figure}
 
\begin{figure}
\centering
\includegraphics[width=0.95\linewidth]{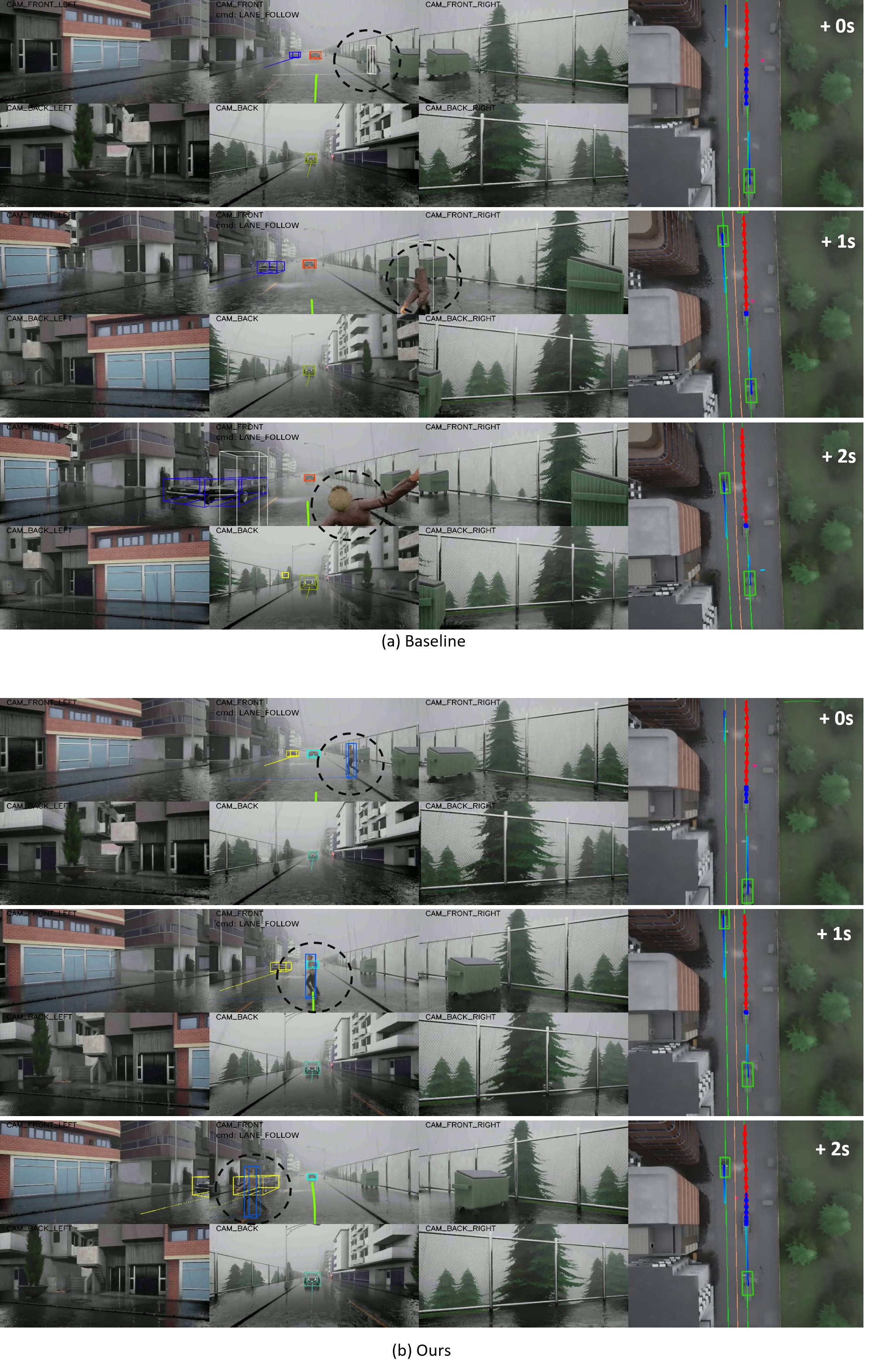} 
\caption{Comparison of Baseline (a) and Ours (b) in a pedestrian cut-in scenario. 
The baseline model fails to avoid the pedestrian, resulting in a collision, whereas our method promptly reacts and avoids the accident. The pedestrian is highlighted with a black dashed circle.}
\label{fig:collision_pred_combined}
\end{figure}

\vspace{-0.15cm}
\paragraph{Effect of Augmentation Probability.} 
Fig.~\ref{fig:ablation_aug} further illustrates the impact of augmentation across different scenarios, showing that performance slightly declines when the augmentation probability exceeds 0.1, as excessive augmentation may encourage overly conservative driving. Overall, all augmented variants substantially outperform the no-augmentation baseline, demonstrating the benefit of our strategy.

\vspace{-0.15cm}
\paragraph{Experimental Reproducibility.}
Due to the inherent stochasticity in the CARLA closed-loop simulator, the results of a single run may slightly. To provide a more comprehensive and reliable reference, we report multiple simulation runs of our base model and compute their average performance in Tab~\ref{tab:multi_run_ourmethod}. Despite these fluctuations, all runs consistently achieve state-of-the-art results, demonstrating the robustness of our approach. This protocol ensures that the reported performance is representative and not an artifact of random variations in the simulation environment.

\section{Visulization}
\label{visulization}
\vspace{-0.2cm}

To better illustrate the effectiveness of our design, we compare our model with the baseline that predicts driving path and trajectory independently (Variant A in Tab.~\ref{tab:module_ablation}) under different challenging interactive scenarios. As shown in \cref{fig:collision_show}, we consider a multi-vehicle intersection scenario where the ego vehicle intends to turn right while yielding to cross traffic. The baseline fails to react to the incoming vehicle (highlighted in purple), resulting in a conflict and eventual collision. In contrast, our method successfully anticipates the cross traffic, waits until it passes, and executes a safe maneuver.

We further evaluate a more critical scenario where a pedestrian suddenly emerges onto the road. As shown in \cref{fig:collision_pred_combined}, the baseline again fails to respond appropriately, leading to a severe safety incident, whereas our method promptly reacts to the pedestrian and avoids the collision. These qualitative results consistently demonstrate the superior safety-aware decision-making capability of our approach under dynamic and unexpected interactions. More visualization results and video demonstrations are provided in the supplementary material.

\section{Limitation}
Recent advances in generative models, such as diffusion- and flow-matching-based approaches, have shown strong potential in producing more realistic trajectory distributions. However, how to incorporate explicit interaction-aware longitudinal reasoning into these generative frameworks remains an open question. In particular, most existing methods still focus on trajectory-level generation without explicitly modeling the coupling between longitudina decisions and surrounding agent dynamics along the driving path. We view integrating our path-conditioned interaction formulation into generative planning paradigms as a promising direction for future work.




\end{document}